\pdfoutput=1

\documentclass[11pt]{article}

\usepackage{multirow}

\usepackage{graphicx}

\usepackage{amsmath}
\usepackage{amsfonts}
\usepackage{amssymb}
\usepackage{threeparttable}
\usepackage[ruled]{algorithm2e}

\usepackage{ACL2023}

\usepackage{times}
\usepackage{latexsym}

\usepackage[T1]{fontenc}

\usepackage[utf8]{inputenc}

\usepackage{microtype}

\usepackage{inconsolata}

\title{Unsupervised Selective Rationalization with Noise Injection}

\author{Adam Storek\\
  Columbia University \\
  \texttt{astorek@cs.columbia.edu} \And Melanie Subbiah\\
  Columbia University \\
  \texttt{m.subbiah@columbia.edu}\And Kathleen McKeown\\
  Columbia University \\
  \texttt{kathy@cs.columbia.edu}
  }
  
\begin{document}
\maketitle
\begin{abstract}
A major issue with using deep learning models in sensitive applications is that they provide no explanation for their output. To address this problem, unsupervised selective rationalization produces rationales alongside predictions by chaining two jointly-trained components, a rationale generator and a predictor. Although this architecture guarantees that the prediction relies solely on the rationale, it does not ensure that the rationale contains a plausible explanation for the prediction. We introduce a novel training technique that effectively limits generation of implausible rationales by injecting noise between the generator and the predictor. Furthermore, we propose a new benchmark for evaluating unsupervised selective rationalization models using movie reviews from existing datasets. We achieve sizeable improvements in rationale plausibility and task accuracy over the state-of-the-art across a variety of tasks, including our new benchmark, while maintaining or improving model faithfulness.\footnote{Code and benchmark are available at \url{https://github.com/adamstorek/noise_injection}.}
\end{abstract}

\section{Introduction}
With the advent of large pre-trained language models like GPT-3 \cite{NEURIPS2020_1457c0d6}, the size and complexity of deep learning models used for natural language processing has dramatically increased. Yet greater performance and complexity can come at the cost of interpretability, masking anything from implementation mistakes to learned bias.

\begin{figure}[hbt!]
    \centering
    \includegraphics[width=\columnwidth]{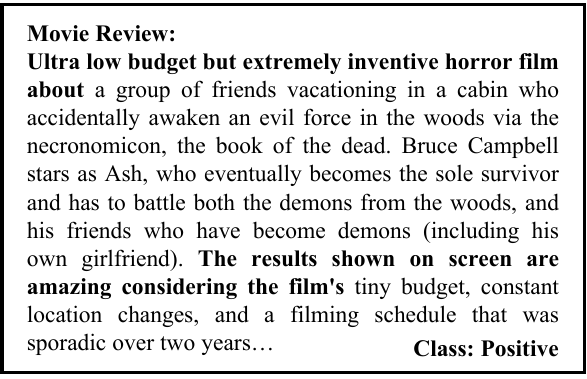}
\caption{\label{fig:x title_page_example} Example of a rationale selected by BERT-A2R + NI (our model) on the USR Movie Review dataset (our benchmark), which asks models to classify movie reviews as positive or negative.}
\end{figure}

A model architecture that justifies its output by providing relevant subsets of input text as a rationale is therefore desirable (see example in Figure \ref{fig:x title_page_example}). The unsupervised selective rationalization architecture as introduced by \newcite{lei_rationalizing_2016} generates rationales alongside predictions by chaining two jointly-trained components, a rationale-generator and a predictor. The generator extracts a rationale: concatenated short and concise spans of the input text that suffice for prediction. The predictor bases its prediction only on this rationale, which encourages \textbf{faithfulness}, meaning how much the rationale reveals what parts of the input were important to the model's prediction. In practice, however, the rationale often isn't \textbf{plausible}, meaning it can't convince a human of the correct prediction, undermining the architecture's interpretability \cite{jacovi-goldberg-2021-aligning, zheng-etal-2022-irrationality}. Using a high-capacity generator can further degrade plausibility \cite{yu-etal-2019-rethinking}.

 To prevent this effect, we introduce a novel training strategy that leverages online noise injection, based on word-level unsupervised data augmentation \cite{50480}. By definition, if the loss-minimizing generator selects an implausible rationale, then the rationale both (a) offers no plausible connection for a human to the target label and (b) locally improves prediction accuracy. This might include communicating via punctuation \cite{yu-etal-2019-rethinking} or subtle input perturbations \cite{garg-ramakrishnan-2020-bae}. Our new approach is to inject noise into the generated rationale during training by probabilistically replacing lower-importance words with noise - random words from the vocabulary - before passing the rationale to the predictor. We observe that this strategy leads to a significant improvement in plausible rationale generation and prediction accuracy without compromising the faithfulness of the architecture. We also show that powerful generators typically interfere with plausible rationale generation but can be effectively deployed when trained with noise injection.
 
To test our approach, we introduce a new benchmark for unsupervised selective rationalization by integrating existing movie review datasets to replace the retracted canonical beer review dataset \cite{mcauley_learning_2012, mcauley_amateurs_2013, lei_rationalizing_2016}.\footnote{The dataset has been retracted at the request of BeerAdvocate and is no longer in use.} We merge a large IMDb movie review dataset \cite{maas-EtAl:2011:ACL-HLT2011} for training and validation and a smaller, rationale-annotated movie review dataset \cite{deyoung_eraser:_2020, zaidan-eisner-2008-modeling, pang-lee-2004-sentimental} for evaluation. We also evaluate our unsupervised approach on the ERASER Movie Review, MultiRC and FEVER tasks \cite{deyoung_eraser:_2020, khashabi-etal-2018-looking, thorne-etal-2018-fever}.\footnote{Licensing information can be found in Appendix \ref{licensing}.}

Our contributions therefore include: 1) characterizing the issue of implausible rationale generation from the perspective of powerful rationale generators, 2) introducing a novel training strategy that limits implausible rationale generation and enables unsupervised selective rationalization models with powerful generators, 3) proposing a new unsupervised rationalization benchmark by repurposing existing movie review datasets, and 4) achieving more plausible rationale generation, with up to a relative 21\% improvement in F1 score and a 7.7 point improvement in IOU-F1 score against the baseline model across a number of tasks.
 
\section{Related Work}
A major challenge with selective rationalization is that discrete selection of rationale tokens is non-differentiable, making training challenging without additional rationale supervision. \newcite{lei_rationalizing_2016} use REINFORCE-style learning \cite{10.1007/BF00992696} to propagate the training signal from the predictor to the generator. \newcite{bastings_interpretable_2019} propose a differentiable approach leveraging the Hard Kumaraswamy Distribution. \newcite{yu-etal-2019-rethinking} strive to improve rationale comprehensiveness. \newcite{chang_invariant_2020} focus on avoiding spuriously correlated rationales. \newcite{yu_understanding_2021} tackle the propensity of selective rationalization models to get stuck in local minima. \newcite{Atanasova_Simonsen_Lioma_Augenstein_2022} use diagnostics-guided training to improve plausibility.

Our work builds on the previous approaches, since we also frame the generator-predictor interaction as a cooperative game and seek to improve plausibility. The previous approaches have, however, introduced additional training objectives \cite{Atanasova_Simonsen_Lioma_Augenstein_2022} or involved incorporating a third adversarial \cite{yu-etal-2019-rethinking} or cooperative \cite{yu_understanding_2021} component. This increases model complexity significantly, leading to more resource-intensive and/or complicated training. Instead, we demonstrate the effectiveness of online noise injection, a considerably more lightweight approach.

An alternative approach is proposed by \newcite{deyoung_eraser:_2020} who assemble a series of datasets with labeled rationales; this enables fully supervised rationale learning. Given rationale-annotated training sets, \newcite{jain_learning_2020} train each model component separately, approaching the accuracy of an entirely black-box model. Although this is a compelling direction, requiring supervision reduces the practical usability of this technique, as many applications lack rationale annotations.

Both unsupervised and supervised selective rationalization approaches generally require a specific token selection strategy to select the output rationale from the generator model \cite{yu_understanding_2021, jain_learning_2020, paranjape-etal-2020-information}.
No previous work that we are aware of, however, has tried to then modify the output rationale before it is input into the predictor. Using online noise injection to enforce prediction stability is therefore a novel approach that adds greater power to the current architectures and can be easily retrofitted.

\section{Implausible Rationale Generation}
Previous work has conceptualized the interaction between the generator and the predictor as a cooperative game \cite{pmlr-v80-chen18j, https://doi.org/10.48550/arxiv.1808.02610, NEURIPS2019_5ad742cd, yu-etal-2019-rethinking, chang_invariant_2020, yu_understanding_2021}. This repeated sequential game consists of two-round stage games. In the first round, the generator accepts an input sequence $X_{1:T}$ and outputs a rationale selection as a binary mask $M_{1:T} \in \mathcal{M}$ where $\mathcal{M}$ represents the set of all masks such that $X_{1:T} \odot M_{1:T}$ satisfies rationale constraints. In the second round, the predictor accepts an input sequence $X_{1:T} \odot M_{1:T}$ and outputs prediction $Y$. The joint objective is to minimize the loss (see Equation \ref{eq joint objective}) based on the generated mask (see Equation \ref{eq generated mask}):
\begin{equation}
M_{1:T} \leftarrow gen(X_{1:T}; \theta_{gen}), M_{1:T} \in \mathcal{M}
\label{eq generated mask}
\end{equation}
\begin{equation}
\min_{\theta_{gen}, \theta_{pre}} \mathcal{L}(pre(X_{1:T} \odot M_{1:T}; \theta_{pre}), \tilde{Y})
\label{eq joint objective}
\end{equation}
For classification, it is customary to minimize the cross-entropy loss $\mathcal{L}_{CE}$. Such a system can be shown to maximize mutual information (MMI) of the rationale with respect to the class label provided sufficient generator and predictor capacity as well as a globally optimal generator \cite{yu_understanding_2021, pmlr-v80-chen18j}:
\begin{equation}
\max_{M_{1:T} \in \mathcal{M}} I(X_{1:T} \odot M_{1:T}; \tilde{Y})
\end{equation}
However, this property does not guarantee rationale plausibility.

First, MMI does not protect against spurious correlations \cite{chang_invariant_2020}. For example, a pleasant taste is not a good explanation for a positive review of a beer's appearance, although the two aspects are strongly correlated.

Second, MMI does not prevent rationale degeneration if the generator and predictor already contain certain biases, for example from pre-training \cite{jacovi-goldberg-2021-aligning}. 

Third, MMI does not prevent rationale degeneration if the generator and predictor are sufficiently powerful to develop a common encoding. \newcite{yu-etal-2019-rethinking} found that providing the generator with a label predicted by a full-input classifier led the generator to develop a communication scheme with the predictor, including a period for positive and a comma for negative examples. \newcite{jacovi-goldberg-2021-aligning} argue that any generator with sufficient capacity to construct a good inner-representation of $Y$ can cause rationale degeneration.

The key underlying cause is that a sufficiently powerful generator is not disincentivized to produce implausible rationales beyond the assumption that generating a plausible rationale should maximize the expected accuracy of the predictor in the current training iteration. However, since the predictor is treated as a black box, this is not guaranteed. On the $i$-th training iteration, the generator greedily selects a binary mask $M_{1:T}$ that minimizes the expected loss:
\begin{equation}
\underset{M_{1:T} \in \mathcal{M}}{\arg\min}\, \mathbb{E} \left[ \mathcal{L}(\widetilde{pre}_i(X_{1:T} \odot M_{1:T}))\right]
\end{equation}
where $\widetilde{pre}_{G, i}$ represents the generator's learned representation of $pre(\cdot; \theta_{pre})$ from its previous experience interacting with the predictor for $i - 1$ iterations in game $G$. As $i$ increases, the generator learns to leverage deficiencies and biases of the predictor that remain hidden to humans, resulting in rationale plausibility degeneration.

\section{Online Noise Injection}
We propose a strategy that disrupts the generator's learned representation of the predictor $\widetilde{pre}_{G, i}$ for all games $G \in \mathcal{G}$, thereby making it harder for the generator to learn to exploit quirks of the predictor. We use online noise injection, which probabilistically perturbs unimportant words in a rationale sequence $X$ of length $T$ (see Algorithm \ref{noise_injection_alg}).

\begin{algorithm}
    \KwIn{$\text{input text }X_{1:T}; \text{binary mask } M_{1:T}$}
    \KwData{$\text{set of documents }\mathcal{D};\text{vocabulary }\mathcal{V}$}
    $R_{1:T} \gets X_{1:T} \odot M_{1:T}$\;
    $R^*_{1:T} \gets R_{1:T}$\;
    \ForAll{$r_i \in R_{1:T}$}{
        $p_i = \text{ProbOfReplacement}_{\mathcal{D}}(r_i)$\;
        $replace \gets Binomial(1, p_i)$\;
        \If{$replace$}{
            $r^*_i \gets \text{SampleFromVocab}_{\mathcal{D}; \mathcal{V}}()$\;
        }
    }
    \KwRet{$\text{perturbed rationale }R^*_{1:T}$}
    \caption{Noise Injection.}
    \label{noise_injection_alg}
\end{algorithm}

If the generator attempts to generate an implausible rationale during training iteration $i$, it strategically includes unimportant words from the input text in the generated rationale, relying on the predictor to pick up on the bias. By subtly perturbing the rationale - replacing the unimportant words - noise injection disrupts this attempt, and the predictor does not respond to the generator-injected bias favorably as expected by the generator. The generator is therefore forced to unlearn/reset its representation $\widetilde{pre}_{G, i}$ of the predictor and reassess its strategy, learning that generating implausible rationales is ineffective. Across any two stages $i, j$ of game $G$, noise injection therefore keeps the learned representations of the predictor more consistent:
\begin{equation}
\forall G \in \mathcal{G}, \forall i,j \in stages(G), \, \widetilde{pre}_{G,i} (\cdot) \approx \widetilde{pre}_{G,j} (\cdot)
\end{equation}

We implement the ProbOfReplacement and SampleFromVocab functions by adapting a strategy that probabilistically replaces words with small TF*IDF, originally proposed for unsupervised data augmentation by \newcite{50480}. We precompute the probability of replacement of each word $w_i \in d$ in each document $d \in \mathcal{D}$ as its normalized TF*IDF score multiplied by the document length and a hyperparameter representing the magnitude of augmentation $p$:
\begin{equation}
\frac{w_{max} - \textit{TF*IDF}(w_i)}{\sum_{w \in d} w_{max} - \textit{TF*IDF}(w)} p |d|
\end{equation}
\begin{equation}
w_{max} = \max_{w \in d} \, \textit{TF*IDF}(w)
\end{equation}

We use these precomputed probabilities to sample which words to replace as shown in Algorithm \ref{noise_injection_alg}. The words are replaced with random words from the vocabulary $\mathcal{V}$. Nonetheless, we also strive to prevent sampling "keywords" from the vocabulary - words that are highly indicative of a label - to avoid confusing the predictor. We compute the sampling probability of $w_i$ as its normalized ATF*IDF, where ATF corresponds to term frequency macro-averaged over $\mathcal{D}$:
\begin{equation}
\frac{w^*_{max} - \textit{ATF*IDF}(w_i)}{\sum_{w \in d} w^*_{max} - \textit{ATF*IDF}(w)}
\end{equation}
\begin{equation}
w^*_{max} = \max_{w \in d} \, \textit{ATF*IDF}(w)
\end{equation}

\section{Model}
\begin{figure*}[htb!]
    \centering
    \includegraphics[width=0.85\textwidth]{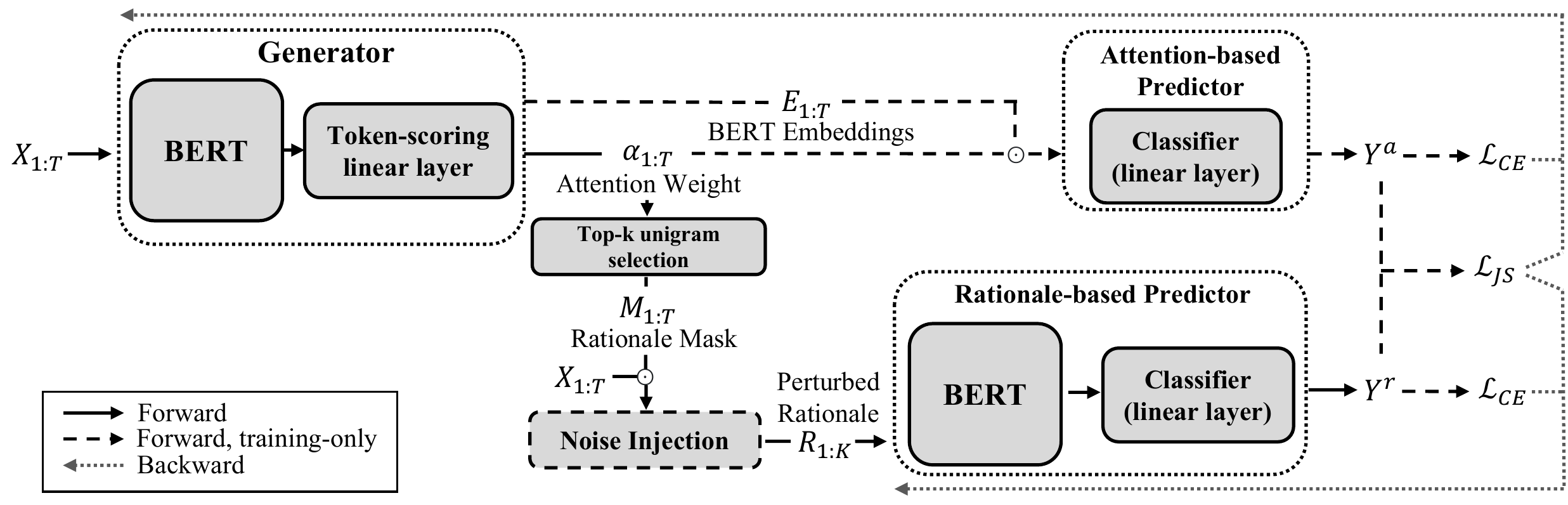}
\caption{BERT-A2R + NI architecture. We replaced the generator's fixed GloVe \cite{pennington-etal-2014-glove} embedding layer used in A2R with BERT-base. The original A2R uses a fixed GloVe embedding layer, GRU \cite{cho-etal-2014-properties}, and a linear classifier pipeline for each predictor. For the attention-based predictor, we remove the GloVe-GRU pipeline and instead reuse the generator's BERT embeddings. For the rationale-based predictor, we replace the GloVe-GRU pipeline with another BERT-base. Both A2R and BERT-A2R feed the masked input text directly into the predictor. To add noise injection during training, we first feed the masked input text into the noise injection component. This component is disabled during evaluation.}
\label{fig:x bert_a2r_ni_model}
\end{figure*}

Our baseline model builds on the A2R architecture by \newcite{yu_understanding_2021} who improve training stability by using an auxiliary predictor connected directly to the generator via an attention layer - this allows for gradients to flow. A2R selects top-$\frac{k}{2}$ bigrams with the highest attention scores from the generator as the rationale and input for the second predictor, with $k$ corresponding to the number of rationale tokens selected as a fraction of the size of the input text. The two components minimize their separate criteria as well as the Jensen-Shannon divergence of their predictions $Y^a$ and $Y^r$ for the attention-based predictor and the rationale-based predictor, respectively. A2R's generator consists of a fixed GloVe \cite{pennington-etal-2014-glove} embedding layer and a linear token scoring layer.

To take full advantage of our noise injection strategy, we replace the limited-capacity generator with BERT \cite{devlin_bert:_2019}. This allows us to use a simpler attention-based predictor than A2R (see Figure \ref{fig:x bert_a2r_ni_model}). To further manifest the efficacy of noise injection, we opt for a top-$k$ unigram selection strategy which offers less regularization compared to a bigram selection strategy. Selecting unigrams is more challenging because it allows the model to select uninformative stopwords like "a" or "the".

Our architecture is shown in Figure \ref{fig:x bert_a2r_ni_model}. Both the selection strategy and the noise injection are model-external and untrained. 
As in \newcite{yu_understanding_2021}, the attention-based (see Equation \ref{eq attention-based loss}) and the rationale-based (see Equation \ref{eq rationale-based loss}) components are trained using identical objectives - minimizing the sum of the cross-entropy loss and the Jensen-Shannon divergence of the two predictors:
\begin{equation}
\mathcal{L}_{a} = \mathcal{L}_{CE}(Y^{a}, \tilde{Y}) +  \lambda JSD(Y^{a}, Y^{r})
\label{eq attention-based loss}
\end{equation}
\begin{equation}
\mathcal{L}_{r} = \mathcal{L}_{CE}(Y^{r}, \tilde{Y}) +  \lambda JSD(Y^{a}, Y^{r})
\label{eq rationale-based loss}
\end{equation}
We refer to our model as BERT-A2R and add +NI when noise injection is used during training.

\section{USR Movie Review Dataset}
\begin{figure}[htb!]
    \centering
    \includegraphics[width=1\columnwidth]{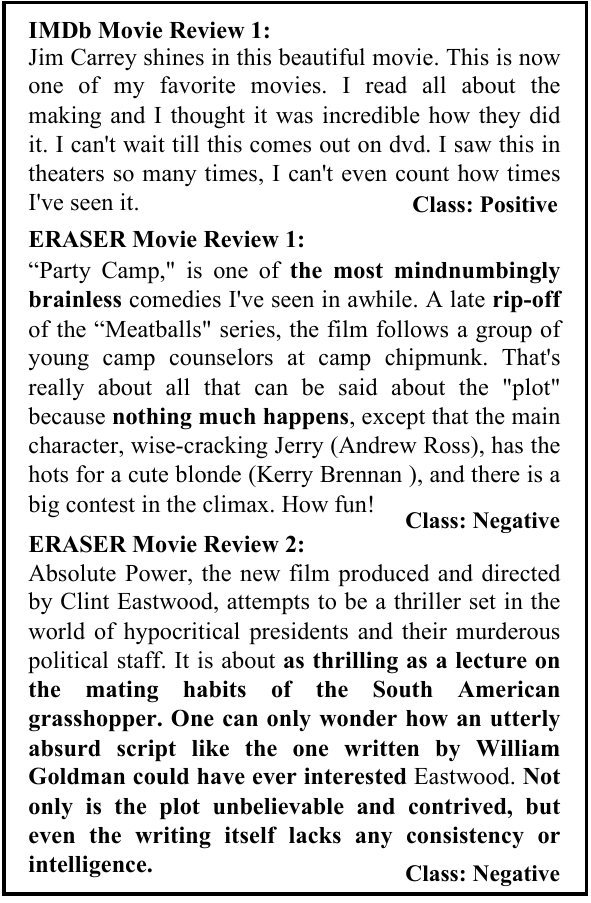}
\caption{Examples from the USR Movie Review Dataset. Note that compared to ERASER reviews, IMDb reviews tend to be shorter; ERASER reviews vary in length dramatically. Furthermore, ERASER rationale annotations are often inconsistent: the rationale for review 1 contains only very short spans, whereas the rationale for review 2 spans a few sentences.}
\label{fig:x dataset_examples}
\end{figure}
Previous work on unsupervised selective rationalization used a decorrelated subset of the BeerAdvocate review dataset \cite{mcauley_learning_2012} as preprocessed by \newcite{lei_rationalizing_2016}. The dataset has recently been removed at the request of BeerAdvocate and is therefore inaccessible to the scientific community. BeerAdvocate reviews consists of 80,000 labeled reviews without rationales for training/validation and $\sim$1,000 labeled reviews with token-level annotated rationales for testing. Alternative datasets either include rationale labels for the entire dataset \cite{deyoung_eraser:_2020} or do not provide rationale labels altogether (e.g. \newcite{maas-EtAl:2011:ACL-HLT2011}).
Moreover, large datasets such as MultiRC or FEVER tend to provide sentence-level rationales compared to BeerAdvocate token-level rationales. We thus repurpose existing movie review datasets to recreate a task similar to beer review, enabling new work on unsupervised selective rationalization to evaluate their performance against models designed for beer review. We merge a smaller ERASER Movie Review dataset \cite{deyoung_eraser:_2020, zaidan-eisner-2008-modeling, pang-lee-2004-sentimental} that has full token-level rationale annotations with the lower-cased Large Movie Review Dataset \cite{maas-EtAl:2011:ACL-HLT2011} which has no rationale annotations.

The movie review task is similar to the binarized beer review task as used in \newcite{NEURIPS2019_5ad742cd, yu-etal-2019-rethinking, chang_invariant_2020, yu_understanding_2021}; both are binary sentiment classification tasks based on English user reviews. However, human rationale annotations of Eraser Movie Review are less coherent and consistent than beer review (see Figure \ref{fig:x dataset_examples}) and lack single-aspect labels comparable to beer review's appearance, aroma, and taste labels. Moreover, movie review annotations tend to be over-complete \cite{yu_understanding_2021}: the same relevant information is often repeated many times in each review. This new task therefore also evaluates previous models' robustness to a subtle distribution shift, an increasingly important consideration for real-world systems.

The reviews from the ERASER Dataset were collected and processed by \newcite{pang-lee-2004-sentimental} from the IMDb archive of the rec.arts.movies.reviews newsgroup, whereas the Large Movie Review Dataset was scraped from the IMDb website by \citet{maas-EtAl:2011:ACL-HLT2011}. In order to avoid overlap between the train and test sets, we looked for similarity by searching for matches between lower-cased, break-tag-free, stop-word-free, lemmatized sentences which spanned at least 5 tokens to avoid generic matches such as "would not recommend" or "great film !". We discovered no overlap between the datasets. We use 40,000 reviews from the Large Movie Review Dataset for training and the remaining 10,000 reviews for validation. We then test our model on the 2,000 annotated examples from ERASER Movie Review.

\section{Experimental setup}
\hspace{0.4cm} \textbf{Metrics} We evaluate generated rationales across several datasets using different metrics that capture faithfulness and plausibility. Faithfulness captures the extent to which the generated rationales truly explain the model's output. For faithfulness, we use comprehensiveness and sufficiency metrics \cite{deyoung_eraser:_2020}. A rationale is \textit{comprehensive} if it extracts all the information contained in the input text that is relevant for prediction and \textit{sufficient} if it contains enough relevant information to make an accurate prediction. The comprehensiveness score measures the difference between the model’s predictions on the entire input text and the input text without the selected rationale (higher is better), whereas the sufficiency score measures the difference between the model’s predictions on the entire input text and just on the rationale (lower is better).

For plausibility, we use standard alignment metrics in reference to the human-annotated rationales: precision, recall, and F1 score as well as IOU-F1 score (referred to as IOU in tables) with partial match threshold 0.1 \cite{deyoung_eraser:_2020, paranjape-etal-2020-information}. We use token-level metrics for Movie Review which offers token-level annotations and sentence-level metrics for MultiRC and FEVER which provide only sentence-level annotations. Finally, we report prediction accuracy for the overall classification task. All results are averaged across 5 random seeds and reported as the mean with standard deviation in parentheses.

\textbf{Implementation}
 Our BERT-A2R models are trained for a maximum of 20 epochs for ERASER Movies and 5 epochs for every other dataset, keeping the checkpoint with the lowest validation loss. All BERT-A2R variants use uncased BERT-base, A2R closeness parameter $\lambda = 0.1$, and the selection strategy of picking the top $k=20\%$ of the highest attention-scoring tokens for movie review or sentences for MultiRC and FEVER. We compute sentence-level scores by taking sentence-level averages of token scores. For optimization, we used Adam \cite{kingma_d.p._adam:_2015} with learning rate 2e-5 and batch size 16. Noise injection level $p$ was set to $0.2$ for USR and ERASER Movie review, $0.3$ for MultiRC, and $0.05$ for FEVER. This was determined based on our hyperparameter search. All of the models were trained on a single machine equipped with a 12-core processor, 64 GB of RAM, and a GPU with 24 GB of VRAM. \footnote{Training details can be found in Appendix \ref{training_details}.}

\section{Results}
\subsection{Does noise injection improve selective rationalization?}
\begin{table}[h]
    \centering
    \begin{tabular}{l||l|l}
        \textbf{Model} & \textbf{Acc.} & \textbf{F1} \\ \hline \hline
        Hard-Kuma \small (\citeyear{bastings_interpretable_2019})& - & 27.0 \\
        BERT Sparse IB \small (\citeyear{paranjape-etal-2020-information})& 84.0 & 27.5 \\
        A2R \small (\citeyear{yu_understanding_2021})& - & 34.9 \\
        BERT-A2R \small (Ours) & 84.0 \small $(2.9)$ & 36.4 \small $(2.8)$ \\ \hline
        BERT-A2R + \textbf{NI} \small (Ours)& \textbf{85.7} \small $(2.7)$ & \textbf{38.6} \small $(0.6)$ \\
    \end{tabular}
    \caption{\label{tab:table eraser movies} Results on ERASER Movie Review (without rationale supervision). +\textbf{NI} indicates using noise injection. We only report Accuracy and F1 to match published results on this benchmark and dashes indicate where the original paper did not publish this metric.}
\end{table}
To compare against previous published results, we trained a BERT-A2R model on the ERASER Movie Review dataset with and without noise injection and compared our numbers to published results from the best unsupervised selective rationalization systems on this benchmark (see Table \ref{tab:table eraser movies}). All models were trained without rationale supervision. We see that our model with noise injection improves on both the classification task accuracy and the rationale F1 score relative to previous systems. Note that noise injection improves the F1 score more than the introduction of BERT to A2R.

\begin{table*}[hbt!]
    \centering
    \begin{threeparttable}[b]
    \begin{tabular}{l|l||l||l|l|l|l||l|l}
          \multicolumn{2}{c||}{}& \multicolumn{1}{c||}{\textbf{Task}} & \multicolumn{4}{c||}{\textbf{Plausibility}} & \multicolumn{2}{c}{\textbf{Faithfulness}} \\ \hline
        \textbf{Dataset} & \textbf{Model} & \textbf{Acc.} & \textbf{P} & \textbf{R} & \textbf{F1} & \textbf{IOU} & \textbf{Com $\uparrow$}  & \textbf{Suf $\downarrow$} \\ \hline
        \multirow{2}{*}{MultiRC} & BA2R & 66.1 \tiny $(1.9)$ & 18.5 \tiny $(1.6)$ & 21.9 \tiny $(2.2)$ & 19.3 \tiny $(1.8)$ & n/a & \textbf{-.01} \tiny $(.01)$ & \textbf{-.02} \tiny $(.02)$ \\
         & BA2R+\textbf{NI} & \textbf{66.4} \tiny $(0.8)$ & \textbf{22.6} \tiny $(1.2)$ & \textbf{26.9} \tiny $(1.8)$ & \textbf{23.8} \tiny $(1.4)$ & n/a & \textbf{-.01} \tiny $(.01)$ & \textbf{-.02} \tiny $(.02)$ \\ \hline
        \multirow{2}{*}{FEVER} & BA2R & \textbf{82.1} \tiny $(3.2)$ & 36.3 \tiny $(0.6)$ & 44.0 \tiny $(0.3)$ & 36.7 \tiny $(0.5)$ & n/a & \textbf{.02} \tiny $(.01)$ & \textbf{-.01} \tiny $(.02)$ \\
         & BA2R+\textbf{NI} & 78.2 \tiny $(1.9)$ & \textbf{39.0} \tiny $(2.5)$ & \textbf{47.2} \tiny $(2.9)$ & \textbf{39.5} \tiny $(2.5)$ & n/a & \textbf{.02} \tiny $(.00)$ & .00 \tiny $(.00)$ \\ \hline
         \multirow{2}{*}{Movies} & BA2R & 84.0 \tiny $(2.9)$ & 36.3 \tiny $(2.8)$ & 36.5 \tiny $(2.8)$ & 36.4 \tiny $(2.8)$ & 30.9 \tiny $(3.9)$ & \textbf{.02} \tiny $(.02)$ & \textbf{-.04} \tiny $(.02)$ \\
         & BA2R+\textbf{NI} & \textbf{85.7} \tiny $(2.7)$ & \textbf{38.5} \tiny $(0.6)$ & \textbf{38.7} \tiny $(0.6)$ & \textbf{38.6} \tiny $(0.6)$ & \textbf{34.4} \tiny $(2.2)$ & \textbf{.05} \tiny $(.02)$ & -.02 \tiny $(.01)$ \\
    \end{tabular}
    \caption{\label{tab:table eraser all} Results on ERASER benchmark datasets. \textbf{P}, \textbf{R}, and \textbf{F1} are sentence-level for MultiRC and FEVER, since they use sentence-level rationale annotations, and token-level for Movie Review, as it uses token-level annotations. \textbf{IOU} is only sensible to use for token-level rationale annotations.}
    \end{threeparttable}
\end{table*}
We then train BERT-A2R models with and without noise injection on the MultiRC and FEVER benchmarks (see Table \ref{tab:table eraser all}) as well as on our new USR Movie Review benchmark (see Table \ref{tab:table generator results}). Again, our noise injection training strategy achieves statistically significant improvements in rationale alignment with human annotations ($p<0.01$ on the MultiRC and USR Movies, $p<0.05$ on the FEVER, and $p<0.1$ on ERASER Movies), achieving up to a relative 21\% improvement in F1 score over our already performant baseline. The plausibility improvement applies for both token-level and sentence-level extraction tasks and across all metrics. Prediction accuracy also improves across all tasks except FEVER. Noise injection also does not seem to have a negative impact on model faithfulness. On ERASER benchmarks, neither comprehensiveness nor sufficiency worsen dramatically, and in the case that one score worsens, the other score tends to remain stable or even improve. On USR movie review, we see an improvement in both faithfulness scores from using noise injection.

\subsection{How does the noise injection level \textit{p} affect model performance?}
\begin{figure}[hbt!]
    \centering
    \includegraphics[width=\columnwidth]{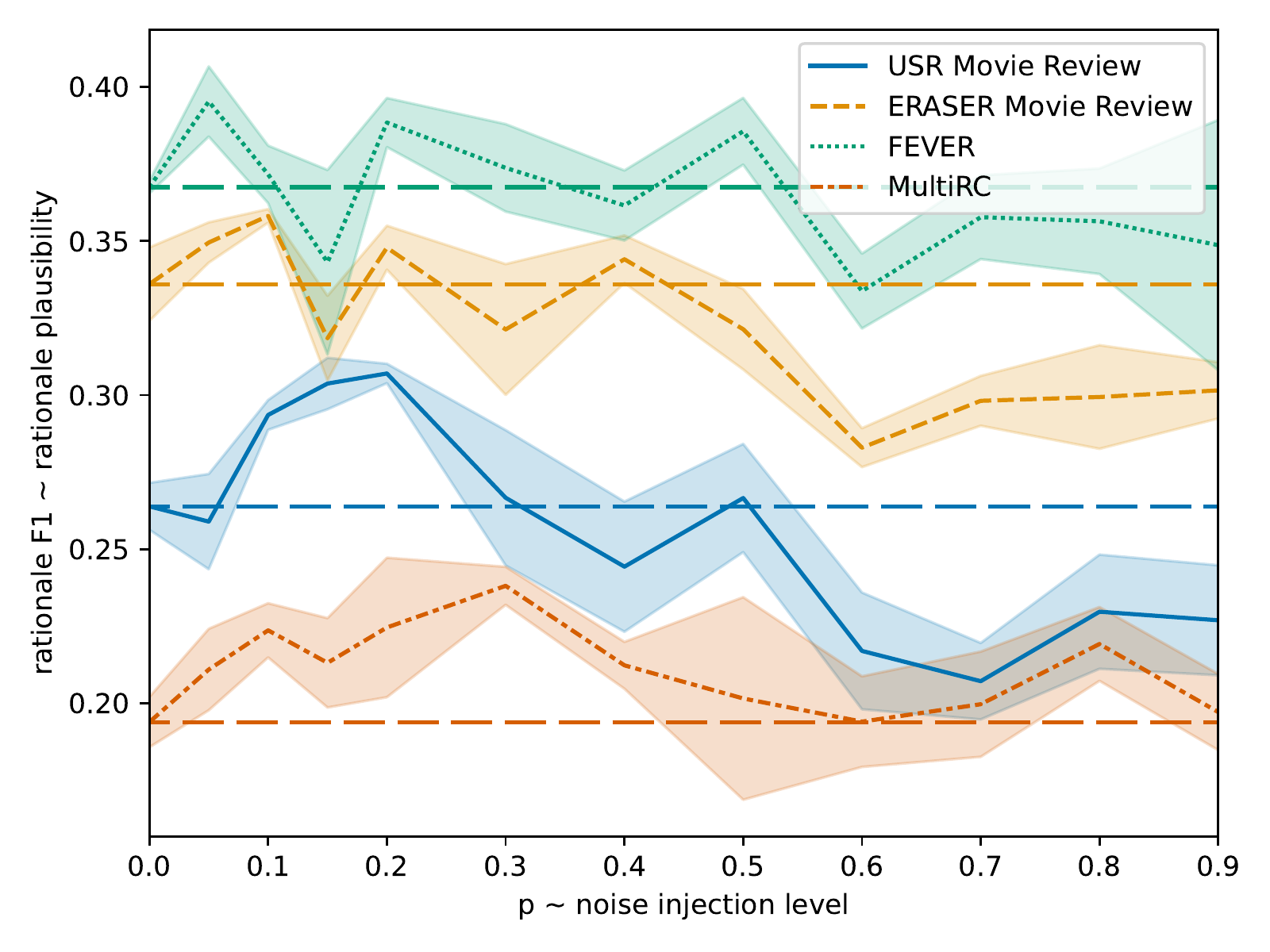}
    \caption{\label{fig:x noise_p_sweep} BERT-A2R + NI rationale F1 on the test set with varying noise injection level $p$. Error bands show $\pm 1$ standard error. Long-dash lines indicate the no noise injection baselines ($p=0$) for each dataset.}
\end{figure}
We train variants of BERT-A2R+NI with different levels of $p$ to examine what noise level is optimal for different datasets (see Figure \ref{fig:x noise_p_sweep}). We average results across 5 seeds but there is still some noise given that the methodology injects noise into the process. It appears that in all cases noise injection seems to degrade performance once $p$ becomes too high as we would expect since too much noise prevents useful signal from getting through. The optimal $p$ varies depending on the task. Rationale alignment performance on FEVER peaks at just $p=0.05$. The optimum for ERASER and USR Movie Review is at $p=0.1$ and $p=0.2$, respectively. The best performance on MultiRC was achieved at $p=0.3$. There are numerous factors that might interact with noise injection to cause this behavior: task-specific demands, sentence vs. token-level rationale annotations, and the suitability of other training parameters. These interactions might be complex, especially with training strategies that dynamically adjust $p$ during training. We leave exploration of these factors for future work.

\subsection{Does noise injection enable the use of powerful high-capacity rationale generators?}
\begin{table*}[hbt]
    \centering
    \begin{threeparttable}[b]
    \begin{tabular}{l||l||l|l|l|l||l|l}
          & \multicolumn{1}{c||}{\textbf{Task}} & \multicolumn{4}{c||}{\textbf{Plausibility}} & \multicolumn{2}{c}{\textbf{Faithfulness}} \\ \hline
        \textbf{Model} & \textbf{Acc.} & \textbf{P} & \textbf{R} & \textbf{F1} & \textbf{IOU} & \textbf{Com $\uparrow$}  & \textbf{Suf $\downarrow$} \\ \hline
        fixed gen. weights & 85.0 \tiny $(0.8)$ & 21.9 \tiny $(0.4)$ & 47.4 \tiny $(0.8)$ & 30.0 \tiny $(0.5)$ & 29.9 \tiny $(0.6)$ & .02 \tiny $(.00)$ & -.02 \tiny $(.00)$ \\
        fixed gen. weights + \textbf{NI} & 85.8 \tiny $(1.1)$ & 22.3 \tiny $(0.4)$ & 48.2 \tiny $(0.9)$ & 30.5 \tiny $(0.6)$ & 30.7 \tiny $(0.8)$ & .03 \tiny $(.01)$ & -.01 \tiny $(.00)$ \\
        tuned gen. weights & 82.4 \tiny $(8.6)$ & 20.2 \tiny $(2.2)$ & 43.7 \tiny $(4.7)$ & 27.6 \tiny $(3.0)$ & 29.1 \tiny $(5.3)$ & .03 \tiny $(.02)$ & -.03 \tiny $(.03)$ \\
        tuned gen. weights + \textbf{NI} & \textbf{87.9} \tiny $(1.8)$ & \textbf{24.4} \tiny $(0.6)$ & \textbf{52.7} \tiny $(1.3)$ & \textbf{33.3} \tiny $(0.8)$ & \textbf{38.4} \tiny $(1.9)$ & \textbf{.04} \tiny $(.01)$ & \textbf{-.04} \tiny $(.02)$ \\
    \end{tabular}
    \caption{\label{tab:table generator results} Results on USR Movie Review using fixed or trainable BERT weights in the BERT-A2R generator.}
    \end{threeparttable}
\end{table*}
For this experiment, we train BERT-A2R with fixed or trainable BERT weights in the generator, with or without noise injection, and evaluate on our new USR Movie Review benchmark (see Table \ref{tab:table generator results}). The version with fixed BERT weights in the generator has much less trainable capacity and cannot learn a task-specific text representation, whereas the generator with trainable BERT weights can potentially learn much better rationales or degrade to implausible rationales. 

We find that the tuned generator trained with noise injection achieves superior performance across all the rationalization metrics without compromising prediction accuracy (2.8 improvement in rationale F1 score and a 7.7 improvement in rationale IOU-F1 score relative to the fixed setting). In contrast, the tuned generator without noise injection training performed the worst in all rationale metrics as well as prediction accuracy. Noise injection with a fixed generator results in a minor improvement in both plausibility metrics and prediction accuracy. We can therefore observe not only that noise injection allows us to leverage the power of a tunable BERT model in the generator that previously would have resulted in performance degradation, but also that the benefits of noise injection are greater with a powerful high-capacity generator model. 

Finally, the addition of noise injection training also slightly improves comprehensiveness for both fixed and tuned generators while improving sufficiency for the tuned generator.
\begin{figure}[hbt!]
    \centering
    \includegraphics[width=\columnwidth]{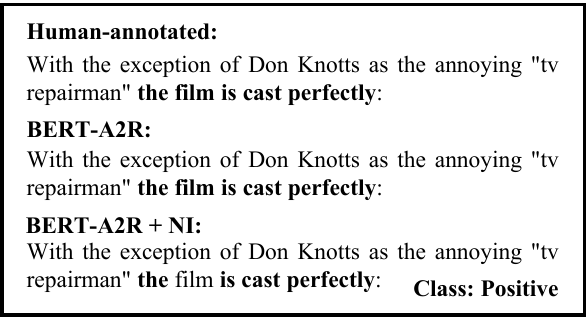}
\caption{\label{fig:x review 1660} An occasional failure case of noise injection training - omitting frequently used words in movie reviews, such  as "film".}
\end{figure}

\subsection{What errors do the models make?}

\begin{figure}[hbt!]
    \centering
    \includegraphics[width=\columnwidth]{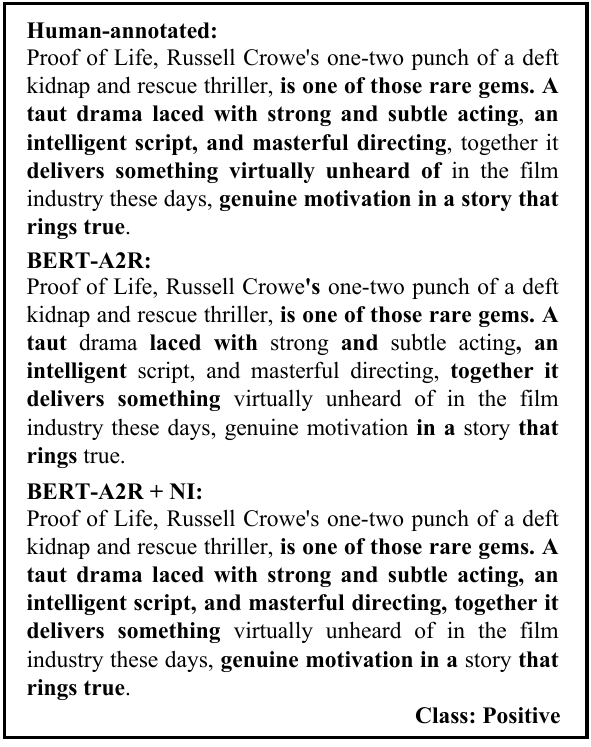}
\caption{\label{fig:x review 762} This review shows the benefits of BERT-A2R + NI's propensity to highlight longer rationale spans where the baseline selects only single words.}
\end{figure}

\begin{figure}[hbt!]
    \centering
    \includegraphics[width=\columnwidth]{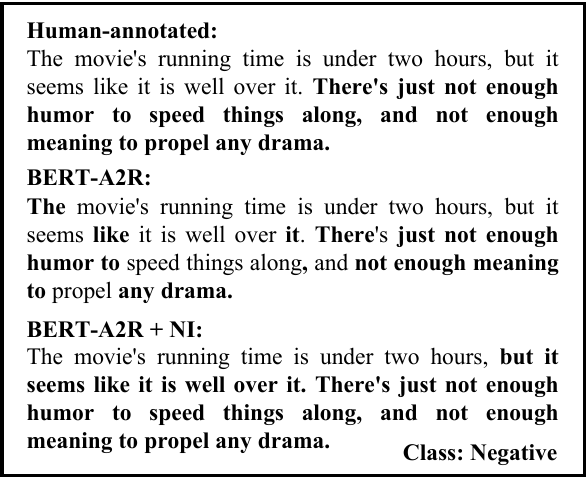}
\caption{\label{fig:x review 124} BERT-A2R + NI produces a more continuous and readable rationale, but it also includes a not-so-relevant part of the previous sentence.}
\end{figure}

For our qualitative analysis we randomly selected 20 reviews to evaluate the effect of adding noise injection to BERT-A2R during training. From this review sample, we include examples that we believe are characteristic for the behavior we observed. First, a BERT-A2R trained with noise injection tends to select longer spans of text as rationales (see Figure \ref{fig:x review 762}, \ref{fig:x review 124}), generally without sacrificing precision compared to the baseline. Selecting continuous rationales greatly improves readability and human-alignment as noted by \newcite{lei_rationalizing_2016}.

We also observed that BERT-A2R + NI occasionally fails to select generic words such as "film" that, nevertheless, form a part of the rationale (see Figure \ref{fig:x review 1660}). This could be a downside to our noise injection strategy, since the model will learn to ignore words with low TF*IDF even though they are relevant in a minority of cases. A potential remedy might be to use task-specific heuristics to generate probability of replacement information instead of the general low TF*IDF strategy. We leave this for future work.

\section*{Conclusion}
In this paper, we investigate a major obstacle of unsupervised selective rationalization frameworks, where the generator has a tendency to learn to generate implausible rationales: rationales that lack a convincing explanation of the correct prediction. We explain the generator's propensity towards degeneration in terms of a flawed incentive structure, characterizing unsupervised selective rationalization as a sequential, repeated cooperative game. Through this lens, we propose a novel training strategy that penalizes implausible rationale generation, thereby realigning the incentive structure with the objective to generate plausible rationales. Using a new benchmark for unsupervised selective rationalization, we show that our noise injection approach is beneficial for training high-capacity generators, outperforming the current state of the art models.

\section*{Limitations}
One of the main limitations of the noise injection training strategy is that statistics used to determine probability of replacement and sampling probability are token-specific. Although this works well on languages with limited morphology such as English, inflected languages like Czech that rely on declension and conjugation might require a lemma-based strategy or a different technique altogether. Furthermore, the model extracts a rationale of fixed length $k$, proportional to the length of the input text. Nevertheless, input text might include more or less information relevant to the class label; a sparsity objective as proposed by \newcite{paranjape-etal-2020-information} could remedy this issue. Lastly, injecting noise during training sometimes leads to more unpredictable training runs.

Additional model limitations are connected to using BERT. Despite its performance and fast training, using BERT limits the scalability to long text due to the 512-token limitation; nevertheless, tasks involving long text might be able to leverage specialized approaches such as \newcite{beltagy_longformer:_2020}. Likewise, BERT renders BERT-A2R about 20 times larger than the GRU-based A2R, requiring greater GPU resources.

The dataset also comes with a few limitations. As \newcite{yu_understanding_2021} note, some reviews contain many clear explanations for the target label, decreasing the need for the generator to include all relevant explanations in the rationale. Similarly, the sparsity of human-annotated rationales can be inconsistent across reviews: as shown in Figure \ref{fig:x dataset_examples}, some rationales include long, generous spans of text that contain irrelevant information, whereas other rationales consist of merely the most important phrases.

\section*{Ethics Statement}
We believe that improving the effectiveness and efficiency of unsupervised selective rationalization in the context of large pre-trained models such as BERT \cite{devlin_bert:_2019} can help uncover and mitigate their learned bias as well as any implementation mistakes. Enabling models to produce plausible faithful rationales increases transparency, improving the end-user's understanding of the model's prediction and allowing AI practitioners to make more informed ethical choices in deploying models.

\section*{Acknowledgments}
This research was conducted with funding from the Defense Advanced Research Projects Agency (DARPA) under Contract No. HR001120C0123. The views, opinions, and findings expressed are those of the authors and do not represent the official views or policies of the U.S. Department of Defense or the U.S. Government.

\bibliography{anthology,custom}
\bibliographystyle{acl_natbib}
\appendix
\section*{Appendix}
\newpage
\section{Licensing} \label{licensing}
\begin{table}[hbt!]
    \begin{tabular}{c||c}
    \textbf{Model} & \textbf{License} \\ \hline \hline
    A2R &  MIT License \\ \hline
    HF BERT-base-uncased & Apache 2.0 \\ \hline
    NLTK "popular" & Apache 2.0
    \end{tabular}
    \caption{\label{tab:table model licensing} Listing of model licenses.}
\end{table}
\begin{table}[hbt!]
    \begin{tabular}{c||c}
    \textbf{Dataset} & \textbf{License} \\ \hline \hline
    FEVER &  Apache License 2.0 \\ \hline
    MultiRC & Apache License 2.0 \\ \hline
    Movies & Apache License 2.0 \\ \hline
    IMDb Movies & None, to our knowledge \\ \hline
    USR Movies & MIT License
    \end{tabular}
    \caption{\label{tab:table dataset licensing} Listing of dataset licenses.}
\end{table}
\section{Training Details} \label{training_details}
Total estimated GPU hours spent on training: 500.
BERT-A2R has 109484547 parameters.
\begin{table}[hbt!]
    \begin{tabular}{c||c|c|c}
    \textbf{Dataset} & \textbf{Train} & \textbf{Val} & \textbf{Test} \\ \hline \hline
    FEVER &  97957 & 6122 & 6111 \\ \hline
    MultiRC & 24029 & 3214 & 4848 \\ \hline
    Movies &  1600 & 200 & 200 \\ \hline
    USR Movies & 40000 & 10000 & 2000
    \end{tabular}
    \caption{\label{tab:table dataset details examples} Dataset details: Number of examples.}
\end{table}

\begin{table}[hbt!]
    \begin{tabular}{c||c|c}
    \textbf{Dataset} & \textbf{Train} & \textbf{Test} \\ \hline \hline
    FEVER &  150 min & 150 s \\ \hline
    MultiRC & 70 min & 90 s \\ \hline
    Movies &  17 min & 15 s \\ \hline
    USR Movies & 110 min & 70 s
    \end{tabular}
    \caption{\label{tab:table dataset details runtime} Dataset details: BERT-A2R runtime.}
\end{table}

\begin{table}[hbt!]
    \begin{tabular}{c||c|c|c|c}
    \textbf{Dataset} & \textbf{LR} & \textbf{BS} & \textbf{\#E} &\textbf{P} \\ \hline \hline
    FEVER & 2e-5 & 16 & 5 & 2 \\ \hline
    MultiRC & 2e-5 & 16 & 5 & n/a \\ \hline
    Movies & 2e-5 & 16 & 20 & 5 \\ \hline
    USR Movies & 2e-5 & 16 (64) & 5 (10) & 2 (n/a)
    \end{tabular}
    \caption{\label{tab:table training details} BERT-A2R Training parameters by dataset. \textbf{LR}, \textbf{BS}, \textbf{\#E} and \textbf{P} stand for learning rate, batch size, number of epochs, and patience. Parameters in parentheses are for fixed BERT generator training.}
\end{table}

\end{document}